\documentclass{article}
\usepackage{ijcai19}

\usepackage{times}
\usepackage{soul}
\usepackage{url}
\usepackage[hidelinks]{hyperref}
\usepackage[utf8]{inputenc}
\usepackage[small]{caption}
\usepackage{graphicx}
\usepackage{amsmath}
\usepackage{booktabs}
\usepackage{algorithm}
\usepackage{algorithmic}
\title{Model soups to increase inference without increasing compute time}

\author{
Charles Dansereau\and
Milo Sobral \and
Maninder Bhogal\And
Mehdi Zalai
\affiliations
Département de génie informatique et génie logiciel, Polytechnique Montréal, Québec, Canada\\
\emails
\{milo.sobral, charles.dansereau, maninder.bhogal, mehdi.zalai\}@polymtl.ca
}

\begin{document}

\maketitle

\begin{abstract}
  In this paper, we compare Model Soups performances on three different models (ResNet, ViT and EfficientNet) using three Soup Recipes (Greedy Soup Sorted, Greedy Soup Random and Uniform soup) from [1], and reproduce the results of the authors. We then introduce a new Soup Recipe called Pruned Soup. Results from the soups were better than the best individual model for the pre-trained vision transformer, but were much worst for the ResNet and the EfficientNet. Our pruned soup performed better than the uniform and greedy soups presented in the original paper. We also discuss the limitations of weight-averaging that were found during the experiments. The code for our model soup library and the experiments with different models can be found \href{https://github.com/milo-sobral/ModelSoup}{here}. 
\end{abstract}

\section{Introduction}

Traditionally, training a deep learning model is done in two steps: first, the model is trained several times by varying the hyper parameters, and then the model with the best performance is chosen before being used. The publication "Model soups: averaging weights of multiple fine-tuned models improves accuracy without increasing inference time" [1] proposes to average the weights of the different models to maximize the use of all the training without wasting learning time. This adds very little memory cost, learning time or inference time. The paper shows that a performance increase is visible on some models and even defines a new state of the art on image classification. The authors present two ways to combine two models to create a soup: greedy soup and uniform soup. In this project, we will first test the Model Souping on a Resnet [3], ViT [4], and EfficientNet [5] to reproduce similar performance improvements. Then, we test a new way to create a soup similar to the pruning of decision trees.

\section{Summary}
\subsection{Prior Work}
Before the paper on model soups, the idea of combining the outputs of multiple models was already explored. In 1996, Leo Breiman published \hyperlink{https://link.springer.com/article/10.1007/BF00058655}{Bagging predictors}, where he presented a method that generates multiple versions of a predictor and uses them to get an aggregated predictor. Then, in 1999, Eric Bauer and Ron Kohavi published \hyperlink{https://link.springer.com/article/10.1023/A:1007515423169}{An Empirical Comparison of Voting Classification Algorithms: Bagging, Boosting, and Variants}, where they reviewed many voting classification algorithms, like Bagging and AdaBoost, and showed interesting results. In 2000, Thomas G. Dietterich published \hyperlink{https://link.springer.com/chapter/10.1007/3-540-45014-9_1}{Ensemble Methods in Machine Learning}, where he explains why ensembles can outperform single classifiers. All of these ensembling techniques increase the accuracy and robustness of models, but have a shortcoming in the computation cost especially when the number of models is large.

\subsection{Original paper}
\begin{figure}[htp]
    \centering
    \includegraphics[width=0.5\textwidth]{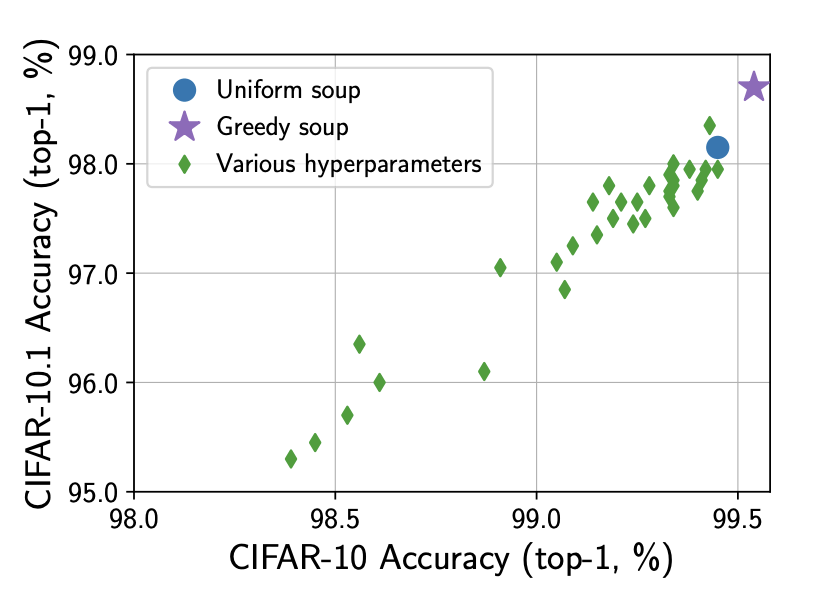}
    \caption{Results of the accuracies obtained by fine-tuning a CLIP ViT-L model on CIFAR-10 in the paper.}
    \label{fig:ModelSoup}
\end{figure}

In the original paper, the authors propose that, instead of selecting the individual fine-tuned model with the highest accuracy on a validation set, averaging the weights of independently fine-tuned models in what they called a \emph{model soup} can outperform the former while requiring no more training and adding no cost of inference. Prior to their paper, weight averaging along a single training trajectory showed a performance improvement in models trained from random initialization [2]. Their approach comes as an extension of the weight averaging concept applied to the context of fine-tuning.\\ \\
The authors explored two main ways to make soups. The first  method is the Uniform soup that simply averages all the models parameters as ${f({x},\frac{1}{k} \sum_{i=1}^k \theta_i})$ where ${\theta_i}$ represents a model found through fine-tuning. The other method is called Greedy soup and is made by first sorting the models by decreasing order of validation accuracy and then by adding each model as an ingredient of the soup if it provides a better performance than the best individual model on the held-out validation set. To use the method of the greedy soup, they consider a neural network denoted by ${f({k},{\theta})}$, where ${\theta}$ represents the parameters that are obtained by fine-tuning models with pre-trained initialization of ${\theta_0}$ and hyperparameter configuration of h as inputs. The configuration of the hyperparameter include the optimizer, data augmentation and the learning rate. The algorithm for the greedy-soup method is illustrated in Algorithm \ref{alg:greedy} \\\\
To evaluate the soup models, the authors fine-tune CLIP, ALIGN and ViT-G models that were all pre- trained. The soups were evaluated on a held out validation set to avoid overfitting the training data.  The results obtained show that the soups outperform the best individual model with the most impressive result being the ViT-G model soup achieving 90.94\% on ImageNet, a new state of the art. It can be seen in figure \ref{fig:ModelSoup} that a Greedy soup definitively results in an improved accuracy. They also show that the model soups are applicable for the fine-tuning of transformer models for text classification. \\\\

\begin{algorithm}[tb]
\caption{Greedy soups}
\label{alg:greedy}
\textbf{Input}: weights of Potential soup ingredients ${\theta_1, ..., \theta_k}$ (optionally sorted in decreasing order of \textbf{ValAcc}($\theta_i$))\\
\begin{algorithmic}[1] 
\STATE soup $\leftarrow \{\}$
\STATE baseline = 0
\FOR{i=1 to k}
\STATE new soup $\leftarrow$ add a model $\theta_i$ to the soup.
\IF {\textbf{ValAcc}(new soup) $>=$ baseline}
\STATE baseline = \textbf{ValAcc}(new soup)
\STATE soup $\leftarrow$ new soup
\ENDIF
\ENDFOR
\STATE \textbf{return} weights of the final soup $\theta_{soup}$
\end{algorithmic}
\end{algorithm}

\subsection{Our approach}
As per the plan submitted for the project, our approach was to first try to replicate the results of the paper on a simpler model. For this, we chose to implement a ResNet model and train it on the CIFAR-100 dataset, which is publicly available. The ResNet model was trained varying the learning rate and weight decay, similarly to the paper, to obtain 36 models with different hyperparameters combinations. We then implemented the Greedy soup and the Uniform soup proposed in the article to evaluate them. To approach the evaluation of the soups the same way the authors of the paper did, a validation set is created by splitting the  original test set of the CIFAR-100 dataset in 2: 5000 examples for the validation set, and 5000 for the new test set. The results are presented and discussed in section 3.\\\\
Following these experiments, we tried to replicate the results obtained by the authors on the same models (ViT-G). Thus, we found pretrained ViT-G models and fine-tuned them with the same hyperparmeters on the CIFAR-100 dataset. We applied the soup-making algorithms and compared our results with the results of the authors in section 3. Furthermore, we proposed a new algorithm to average the weighs of the models in a way similar to pruning in decision trees, which we called "Pruned Soups". The pseudo-code for the algorithm can be found in Algorithm \ref{alg:algorithm}.\\\\
The third model's approach consisted to implement an EfficientNet [6] model  as it is said to achieve a better accuracy and efficiency than previous ConvNets such as Resnet [5]. The model was trained using the same hyperparameters on the CIFAR-100 dataset. The implementation was taken from a code from Github submitted for the paper "EfficientNet: Rethinking Model Scaling for Convolutional Neural Networks". You can find the source code  \href{https://github.com/aladdinpersson/Machine-Learning-Collection/blob/master/ML/Pytorch/CNN_architectures/pytorch_efficientnet.py}{here}. This model was used as it is focused on balancing the dimensions of the network width, depth and resolution by scaling each aspect with constant ratio.

\begin{figure}[htp]
    \centering
    \includegraphics[width=0.5\textwidth]{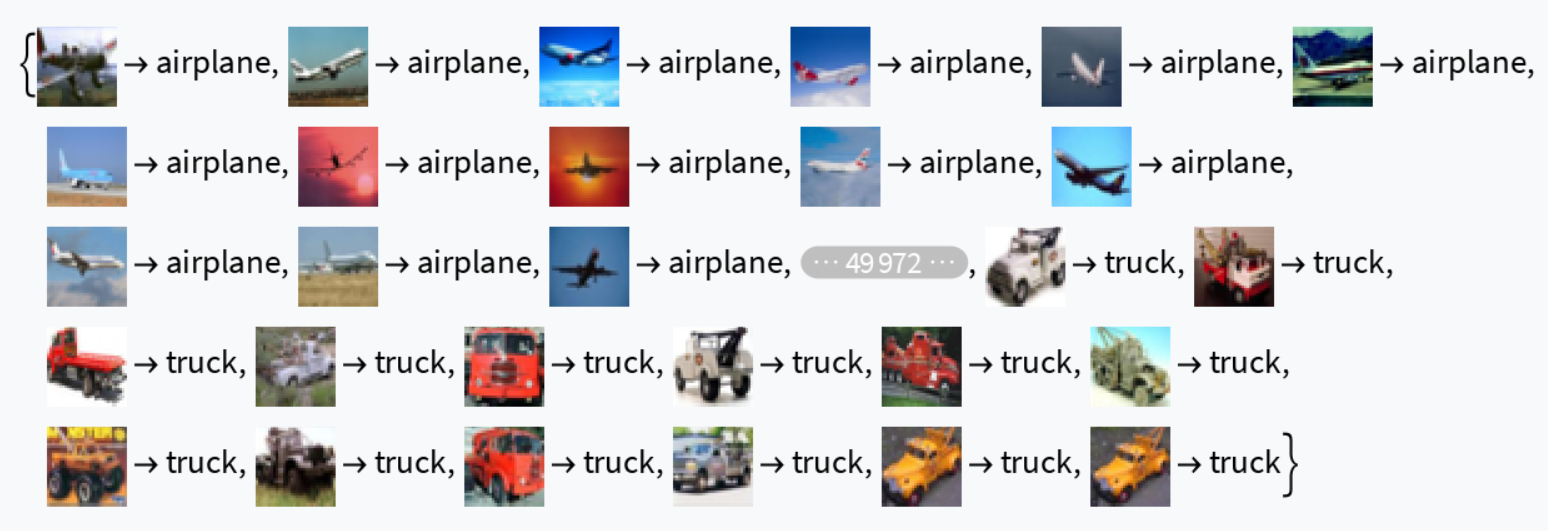}
    \caption{CIFAR-100 dataset on which all our models were trained.}
    \label{fig:cifar-100}
\end{figure}

\begin{algorithm}[tb]
\caption{Pruned soups}
\label{alg:algorithm}
\textbf{Input}: weights of Potential soup ingredients ${\theta_1, ..., \theta_k}$ (optionally sorted in decreasing order of \textbf{ValAcc}($\theta_i$))\\
\textbf{Parameter}: numbers of passes (N)\\
\begin{algorithmic}[1] 
\STATE soup = $\frac{1}{k} \sum_{i=1}^k \theta_i$
\STATE baseline = \textbf{ValAcc}(averaged weights).
\FOR{pass=1 to N}
\FOR{i=1 to k}
\STATE new soup $\leftarrow$ Remove a model $\theta_i$ from the soup.
\IF {\textbf{ValAcc}(new soup) $>=$ baseline}
\STATE baseline = \textbf{ValAcc}(new soup)
\STATE soup $\leftarrow$ new soup
\ENDIF
\ENDFOR
\ENDFOR
\STATE \textbf{return} weights of the final soup $\theta_{soup}$
\end{algorithmic}
\end{algorithm}

\subsection{Implementation Details}

The model soup code has two main components:
\begin{itemize}
    \item The \emph{generate\_soup.py} script which contains the main \emph{make\_soup()} function which uses pytorch trained models and an evaluator to generate different kinds of soups.
    \item An evaluator interface, which must be implemented by the Evaluator object used to generate the soups. An example of an Evaluator Class can be found in \emph{model\_evaluators/cifar\_eval}. 
\end{itemize}
The \emph{Evaluator} provides the metric upon which the classification of the models and the selection of ingredients for the soup will be made. The code is designed to always maximize that metric so we must be careful when using metrics like loss that are designed to be minimized. The \emph{Evaluator} is responsible for loading the desired dataset and iterating over it. The soup code is designed to use two separate sets: one for initial ordering of the models and for selecting ingredients and one for computing final performance. This distinction is critical to avoid overfitting our soups to one single dataset.\\\\

\section{Methods}
\subsection{Experimental setup}
This section represents our experimental setup and the results on CIFAR-100 using three different approaches. The models are trained using similar settings : a stochastic gradient descent optimizer with a 0.9 value for the momentum, and with a weight decay [1e-5, 2e-5, 5e-5, 1e-4, 2e-4, 4e-4] and learning rate [0.01, 0.02, 0.05, 0.1, 0.2, 0.4] varying for each model.  Each model is trained for 10 to 15 epochs with a total batch size of 256.  Models were trained at first over 36 different combinations of hyperparameters, from which the accuracy of the best model was noted.  To create the soups for each model, a \href{https://colab.research.google.com/drive/1yyRSK9x35gErpMy_LQjB4ULR8GagSVVJ?usp=sharing}{Colab notebook} was used to select the type of model to evaluate, the number of runs, the number of ingredients, the type of strategy and the soup making method.  To evaluate the performances of the soups, the test set was separated to create a separated validation set. The user-friendly notebook allows to import the functions easily from Github and test any model, in our case: ResNet, ViT or EfficientNet. Results were obtained for sorted and random Greedy soups, Uniform soups, and sorted and random Pruned soups.

\subsection{Results}

\begin{table}[h]
\centering
\begin{tabular}{lrr}  
\toprule
Method  &  Acc. (\%) & Ingredients (avg)\\
\midrule
Best individual model & 50.3     & - \\
Uniform soup          & 32.22  & 22      \\
Greedy soup (random)  & 51.06  & 5.1       \\
Greedy soup (sorted)  & 51.76  & 5      \\
Pruned soup (random)  & 52.04  & 3.2      \\
Pruned soup (Sorted)  & 52.1  & 3      \\
\bottomrule
\end{tabular}
\caption{Accuracies of the different methods accross our experiments on the CIFAR-100 dataset for the ViT models.}
\label{tab:booktabs}
\end{table}

\begin{figure}[h]
\centering
\includegraphics[width=0.5\textwidth]{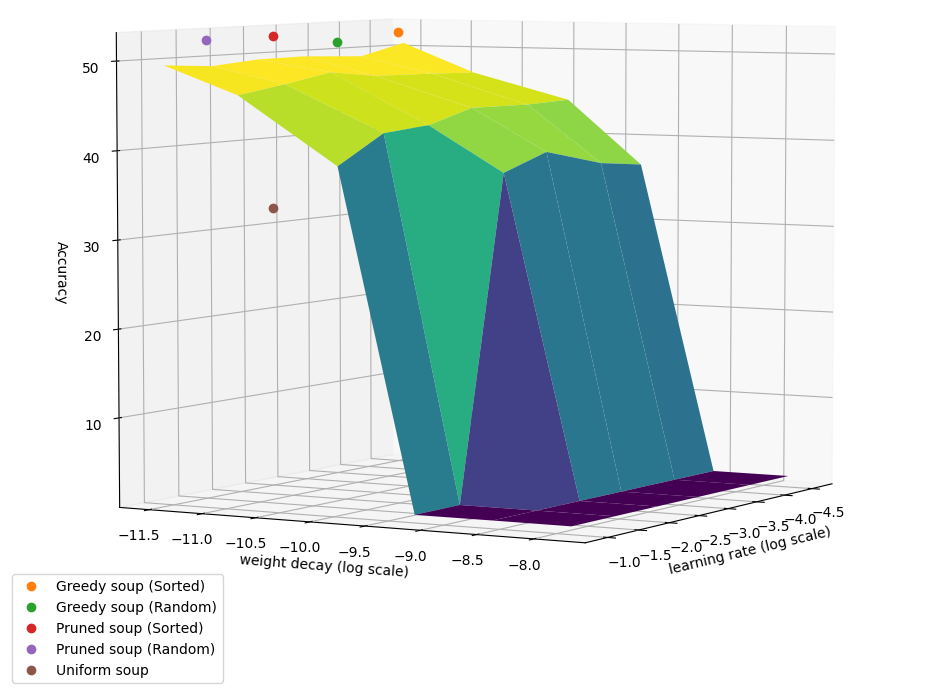}
\caption{Comparison of the performance obtained by the different soups made and the individual models from the grid search for the ViT models.}
\label{surface}
\end{figure}

As we can see on Table \ref{tab:booktabs}, the Greedy soups achieved a better performance than the best fine-tuned model, with an accuracy of 51\% while picking models randomly, and 51.75\% when the models were sorted in decreasing accuracy. This is about 1\% above the performance of the best model (50.3\%) which is comparable to the results obtained in the paper [1]. We can also see that the proposed method (Pruned soups) achieves even higher performance, with an accuracy of 52\% while picking models randomly, and 52.1\% when the models were sorted in decreasing accuracy. However, it is important to note that the standard deviation of these results are in the order of 0.3\%, so more experiments would be needed to conclude if the pruned soups perform better in average than the greedy soups. Another observation that can be made is that the pruned soups used fewer ingredients in average (around 3) than the greedy soups (around 5). Finally, contrary to the paper, the uniform soup has a much lower accuracy than the best model. This can be explained by the fact that some models included in the uniform soup had poor performance, as shown by the surface plot of the grid search in Figure \ref{surface}.

\begin{table}[h]
\centering
\begin{tabular}{lrr}  
\toprule
Method (ResNets)  &  Acc. (\%) & Ingredients (avg)\\
\midrule
Best individual model & 65.78     & - \\
Uniform soup          & 1.08  & 33      \\
Greedy soup (random)  & 65.78  & 1       \\
Greedy soup (sorted)  & 65.78  & 1      \\
Pruned soup (random)  & 1.19  & 6.2      \\
Pruned soup (Sorted)  & 1.19  & 6.2      \\
\midrule
Method (EfficientNets)  &  Acc. (\%) & Ingredients (avg)\\
\midrule
Best individual model & 40.12     & - \\
Uniform soup          & 0.98  & 36      \\
Greedy soup (random)  & 40.12  & 1       \\
Greedy soup (sorted)  & 40.12  & 1      \\
Pruned soup (random)  & 1.12  & 5.1      \\
Pruned soup (Sorted)  & 1.12  & 5.1      \\
\bottomrule
\end{tabular}
\caption{Accuracy of the different methods accross our experiments on the CIFAR-100 dataset for the ResNets and EfficientNets models.}
\label{tab:resnet}
\end{table}

When looking at the ResNets and EfficientNets models, we can instead see that the performance of the soups dramatically decreases in comparison to the best model, as shown in Table \ref{tab:resnet}. In fact, their accuracy is close to 1\%, which means that they do not do better than a model that would completely randomly guess the class of the image, since the CIFAR-100 dataset has 100 possible classes. Further, we can observe that the only soups that still can generalize on the test set are the greedy soups containing only 1 ingredient, which means it is the same as picking the best model or a random model, according to the strategy of the soup.  Therefore, it can be seen that the architectures of these models are not well adjusted for the creation of soups. 

\subsection{Ablation study}

\begin{figure}[h]
\centering
\includegraphics[width=0.5\textwidth]{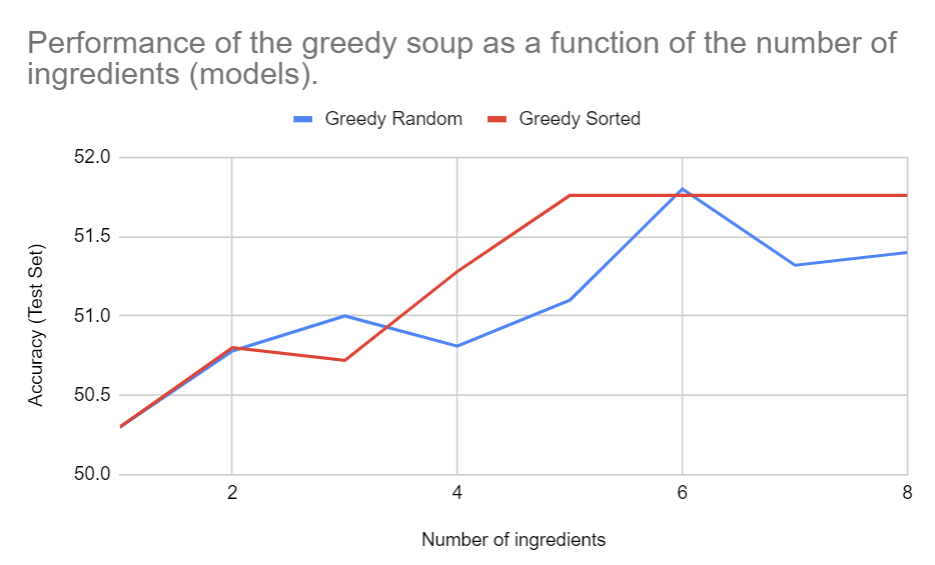}
\caption{Greedy Random and Greedy Sorted comparison}
\label{greedyRdmVsSorted}
\end{figure}

\begin{figure}[h]
\centering
\includegraphics[width=0.5\textwidth]{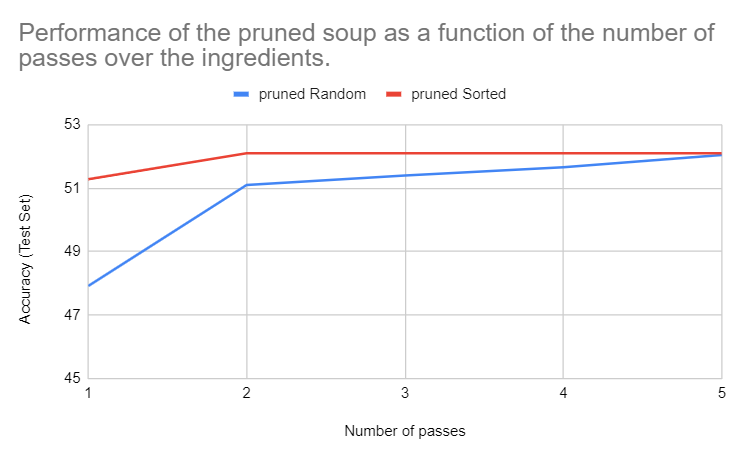}
\caption{Pruned Random and Pruned Sorted comparison}
\label{prunedRdmVsSorted}
\end{figure}

\begin{figure}[h]
\centering
\includegraphics[width=0.5\textwidth]{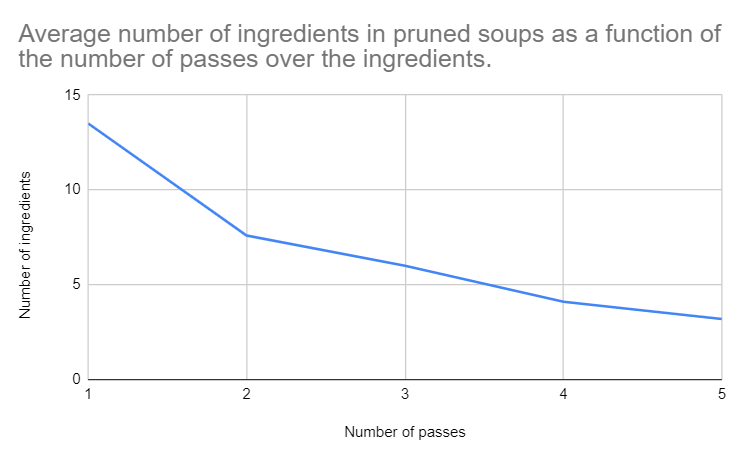}
\caption{Average number of ingredients in pruned soups as a function of the number of passes over the ingredients}
\label{prunedPassesIngredients}
\end{figure}

We illustrate the impact of the number of ingredients on the performances of the greedy soup in figure \ref{greedyRdmVsSorted}. An increase in ingredients  (up to 5) provides a better final model over the best individual model with an improvement of the accuracy by around 1\%.  The greedy soups studied in the paper also select 5 models for CLIP ViT-B/32 and ALIGN EfficientNet-L2 that provides better results by 0.7 and 0.5 percentage points. This applies to both strategies (considering the models to add randomly versus in decreasing order of accuracy). It can also be noted that when increasing the maximum number of ingredients beyond 5, the sorted greedy soup has the same performance, as adding any other model results in a decrease in accuracy. Similarly, the random greedy soup gets diminishing returns and oscillates around 51.5\% accuracy. The oscillation can be explained by the variance in the order in which the models are considered to be added in the soup. To mitigate that, the results were averaged over 10 runs.\\\\
With the help of figure \ref{prunedRdmVsSorted}, we can see that an increase in the number of passes increases the accuracy of both random and sorted pruned soups. However, the sorted pruned soup quickly stabilizes at a 52.1\% accuracy with a number of passes greater or equal to 2. For the random pruned soup, we have that the accuracy increases at every increase of the number of passes finally reaching similar performance to the sorted pruned soup at 5 passes with an accuracy of 52.04\%. This means the number of passes can be increased until the optimal combination of models for the soup is found, and removing any model from the soup hinders its ability to generalize.\\\\
As we can see in figure \ref{prunedPassesIngredients}, an increase in the number of passes has a decreasing impact on the average number of ingredients in pruned soups. Starting at an average of 13.5 ingredients for a single pass, that number decreases exponentially, with a smaller decrease at every increase in the number of passes, to reach a much lower average of 3.2 ingredients, which was the best configuration with the set of trained models that was used. \\\\
Similar to what the paper mentioned, a sorted strategy gives a slightly better accuracy than a randomly ordered strategy. We can see in figures \ref{surface} and \ref{greedyRdmVsSorted} that a better accuracy is achieved with sorted soup models.\\\\\\

\section{Discussion}

\subsection{Analysis}
From the results of the vision transformer, model soups produce a better model by averaging the weights. However, lower results were obtained for the uniform soups: the ViT model obtained an accuracy of 17\% below the best individual model.  Indeed, in the paper, the uniform soup model only achieves a better accuracy than the best individual model when all individual models achieve high accuracy, hence why Greedy soups perform better as they exclude models having a lower accuracy while averaging the weights. This means an hypothesis for a new criteria can be made: the models considered for the uniform soup must have an accuracy in the order of the best model, similar to a set of fine-tuned models with the same local minimums, with small variations. In that sense, if the weights of the models converge to different local minimums, averaging them will move the solution in an arbitrary place, which is not guaranteed at all to be in a local minimum. This was observed for the ResNets and EfficientNets models.\\\\
In order to apply different approaches to the soups, it is important to further analyze and understand the limitations of model soups on ResNets and EfficientNets as the averaging weights method did not work. As the authors mentioned in the paper, the use of pre-trained models is essential to the use of soups for the fine-tuning of hyperparameters. We showed that a simple average of learned parameters of non-pre-trained models made the performance much worse, performing in a similar range to random guesses. In other words, the concept of soups works well for the fine-tuning of pre-trained models, but not for the trained-from-scratch models. In fact, the soups work well when the models have been tuned independently from the same initialization as they have the same error landscape. Averaging the model weights will then result in a coherent performance only when we have this similar error landscape between models. This leads to a limitation that was not discussed in the original paper: the models used to make soups not only need to be pre-trained, but the model in itself will need to converge to the same error landscape when fine-tuned, so the soup methods cannot be applied to all model architectures. For instance, the EfficientNets used for our experiments where in fact pre-trained, but the fine-tuning with different hyperparameters lead to different local minima for the weights, and averaging any model with another in this situation destroys any capacity of the soup to generalize.

\subsection{Future Work}

Although model soups show some promise on some problems and have been shown to provide state of the art results for some problems, a clearer understanding of their limitations is important. In this work, we show some limitations of model soups such as the need for all models to be in the same error landscape. A comprehensive study of all models where model soups might or might not perform would be critical to understand the different limitations of model soups. 
\\\\
One relevant hypothesis is that batch normalization is causing the fine-tuned models to end up in different error landscapes as it introduces some randomness to the weights of the model. Verifying this hypothesis would require some more experimentation and would lead to a much deeper understanding of the limitations of model soups. Another hypothesis is that the different error landscapes might arise as a result of the optimizer used during fine-tuning of the models. To verify that hypothesis, more experiments could be done with different optimizers to see if this factor has an influence.  

\section{Conclusion}

This work shows that Model Soups are promising when it comes to making the most out of the training time of your models and is a step forward to getting to smarter hyperparameter tuning. However, we show that the shortcomings of model soups are still not well known and understood. In order to further develop the concept of model soups and integrate it in modern deep learning pipelines, these limitations will need to be understood and, if possible, fixed completely. Once they are well understood, model soups could very well become an integral part of modern deep learning pipelines as their benefits towards training efficiency and performance improvements cannot be undermined. 

\newpage
\section{Bibliography}
\medskip
[1] Wortsman, M., Ilharco, G., Yitzhak Gadre Rebecca Roelofs, S., Gontijo-Lopes, R., Morcos, A. S., Namkoong Ali Farhadi, H., Carmon, Y., Kornblith, S., and Schmidt, L. (n.d.). Model soups: averaging weights of multiple fine-tuned models improves accuracy without increasing inference time.
\newline
\newline
[2] Izmailov, P.,  Podoprikhin, D., Garipov , T., Vetrov, D.,  Gordon Wilson, A.  2018. Averaging Weights Leads to Wider Optima and Better Generalization.
\newline
\newline
[3] He, K., Zhang, X., Ren, S., and Sun, J. (2015). Deep Residual Learning for Image Recognition.
\newline
\newline
[4] Ross Wightman. PyTorch Image Models. https://github.com/rwightman/pytorch-image-models. Retrieved April 13, 2022
\newline
\newline
[5] Tan, M. (2021, April 1). EfficientNet: Rethinking Model Scaling for Convolutional Neural Networks. https://arxiv.org/pdf/1905.11946.pdf. Retrieved April 18, 2022.
\newline
\newline
[6] Aladdin Persson. CNN Architectures. https://github.com/aladdinpersson/Machine-Learning-Collection. Retrieved April 18, 2022.
\newline
\newline

\end{document}